\documentclass[10pt,journal,twoside]{IEEEtran}

\ifCLASSOPTIONcompsoc
\usepackage[nocompress]{cite}
\else
\usepackage{cite}
\fi

\ifCLASSINFOpdf
\else
\fi

\usepackage{amsmath}
\usepackage{bbm}
\usepackage{enumerate}
\usepackage{booktabs}
\usepackage{multirow}
\usepackage{graphicx}
\usepackage{subfigure}
\usepackage{color}
\usepackage{algorithm}
\usepackage{algorithmic}
\usepackage{hyperref}[colorlinks, linkcolor=black]

\usepackage{graphicx}
\usepackage{latexsym}

\usepackage{times}
\usepackage{soul}
\usepackage{url}

\usepackage{amsthm}
\usepackage{booktabs}

\usepackage[switch]{lineno}
\usepackage{bbding}

\usepackage{graphics,color,multirow,subfigure}
\usepackage{makecell}

\usepackage{amsfonts}

\begin{document}
	%
	\title{Mitigating the Structural Bias in Graph \\ Adversarial Defenses}
	%
	%
	%
	%
	\author{Junyuan Fang, \IEEEmembership{Student Member, IEEE}, Huimin Liu, Han Yang, Jiajing Wu, \IEEEmembership{Senior Member, IEEE},\\ 
		Zibin Zheng, \IEEEmembership{Fellow, IEEE}, and Chi K. Tse, \IEEEmembership{Fellow, IEEE}
		\thanks{Manuscript received xxx. \emph{(Corresponding Author: Jiajing Wu.)}}
		\IEEEcompsocitemizethanks{ 
			\IEEEcompsocthanksitem{Junyuan Fang and Chi K. Tse are with the Department of Electrical Engineering, City University of Hong Kong, Hong Kong SAR, China. (Email: junyufang2-c@my.cityu.edu.hk; chitse@cityu.edu.hk)}
			\IEEEcompsocthanksitem{Huimin Liu and Han Yang are with the School of Computer Science and Engineering, Sun Yat-sen University, Guangzhou 510006, China. (Email: liuhm59@mail2.sysu.edu.cn; yangh396@mail2.sysu.edu.cn)}
			\IEEEcompsocthanksitem{Jiajing Wu and Zibin Zheng are with the School of Software Engineering, Sun Yat-sen University, Zhuhai 519082, China. (Email: wujiajing@mail.sysu.edu.cn; zhzibin@mail.sysu.edu.cn)}}
		\thanks{Digital Identifier: IEEE.xxx.xxx.xxxx.xxxxx.}
	}
	%
	%

	\markboth{IEEE Transactions on xxx,~Vol.~XX, No.~XX, XX~2024}%
	{Fang \MakeLowercase{\textit{et al.}}: Mitigating the Structural Bias in Graph Adversarial Defenses}
	
	\maketitle

	\begin{abstract}		
		In recent years, graph neural networks (GNNs) have shown great potential in addressing various graph structure-related downstream tasks. However, recent studies have found that current GNNs are susceptible to malicious adversarial attacks. Given the inevitable presence of adversarial attacks in the real world, a variety of defense methods have been proposed to counter these attacks and enhance the robustness of GNNs. Despite the commendable performance of these defense methods, we have observed that they tend to exhibit a structural bias in terms of their defense capability on nodes with low degree (i.e., tail nodes), which is similar to the structural bias of traditional GNNs on nodes with low degree in the clean graph. Therefore, in this work, we propose a defense strategy by including hetero-homo augmented graph construction, $k$NN augmented graph construction, and multi-view node-wise attention modules to mitigate the structural bias of GNNs against adversarial attacks. Notably, the hetero-homo augmented graph consists of removing heterophilic links (i.e., links connecting nodes with dissimilar features) globally and adding homophilic links (i.e., links connecting nodes with similar features) for nodes with low degree. To further enhance the defense capability, an attention mechanism is adopted to adaptively combine the representations from the above two kinds of graph views. We conduct extensive experiments to demonstrate the defense and debiasing effect of the proposed strategy on benchmark datasets.
	\end{abstract}
	

	\maketitle


	%
	\IEEEpeerreviewmaketitle

	\section{Introduction}\label{sec:intro}
	\IEEEPARstart{B}{y} leveraging the strong learning capability of the message-passing mechanism, i.e., neighborhood aggregations, graph neural networks (GNNs) have achieved great success in a variety of graph prediction tasks, such as node classification, link prediction, graph clustering, etc. \cite{wu2020comprehensive,zhou2020graph,zhang2020deep,jiang2021graph}. Specifically, besides ego features, each node in the graph can further utilize the information from its neighbors by aggregating the features of the neighboring nodes. This success underscores the vast potential of GNNs in fields such as social network analysis, recommendation systems, and bioinformatics, demonstrating their promising prospects for future applications.

	Despite the impressive performance of GNNs on the aforementioned prediction tasks, they have been pointed out to be vulnerable against some small unnoticeable perturbations, which are also called graph adversarial attacks \cite{jin2020adversarial,xu2020adversarial,chen2020survey}. In a typical graph adversarial attack scenario, attackers attempt to inject some noise into the original clean graph, such as adding/removing links, modifying node features, injecting fake nodes, etc., aiming to mislead the downstream predictions of GNNs. Over a recent period, a number of adversarial attack methods have been proposed to evaluate the robustness of current GNNs from diverse perspectives \cite{zugner2018adversarial,wu2019adversarial,dai2018adversarial,chen2018fast,tao2021single}.
	
	As the sophistication of attacks advances, some defense strategies have also been proposed to counter and mitigate these evolving threats 	\cite{wu2019adversarial,entezari2020all,zhu2019robust,ijcai2021p310,deng2023batch}. Unlike in adversarial attacks, defenders aim to eliminate the negative influence of perturbations injected by attackers through methods such as graph purification, robust GNN aggregators, etc. By leveraging these mechanisms, the robustness of GNNs has been significantly improved.

	However, similar to the long tail issue prevalent in traditional GNNs where nodes with high degree typically achieve superior prediction accuracy than nodes with low degree \cite{liu2020towards}, current defense methods also confront the same structural bias during prediction. Details are presented in Section \ref{sec:emp}. In general, many real-world networks demonstrate a scale-free degree distribution, namely, a large proportion of nodes with low degree while a small proportion of nodes with high degree. Therefore, since the performance of GNNs is strongly related to the effect of neighborhood information, the neighborhood aggregations of nodes with low degree tend to be somewhat incomplete due to insufficient neighborhoods. These observations disclose that current defense GNNs tend to have a bias in terms of structure, which could not give the same recovery effect to nodes with high degree and low degree. On the other hand, although a series of enhanced GNNs have been proposed to mitigate the structural bias, few of them take into account the more challenging scenarios involving adversarial attacks \cite{tang2020investigating,liu2021tail,yun2022lte4g,kang2022rawlsgcn,liu2023generalized}.

	Therefore, in this work, we propose a simple yet effective GNN framework, De2GNN, to  {\em de}fend against possible adversarial attacks and {\em de}bias the structural bias simultaneously. Specifically, the proposed De2GNN includes the following three modules, namely the hetero-homo augmented graph construction, $k$NN augmented graph construction, and multi-view node-wise attention mechanism. We first detect possible heterophilic links in the original (possibly perturbed) graph by removing heterophilic links, aiming to mitigate the negative influence of possible adversarial perturbations. Then, we enrich the neighborhood information of nodes with low degree by adding potential neighbors based on the guidance from a surrogate GNN model via adding homophilic links for nodes with low degree. From these two steps, we can obtain a relatively pure and informative graph, labeled as the hetero-homo augmented graph. Moreover, we construct a new augmented $k$NN graph based on the ego features of each node, which is attack-agnostic. Finally, we utilize two separate GNNs to obtain the information of each node from the above two graph views and then adaptively combine the corresponding information through a node-wise attention mechanism. Comprehensive experiments have been conducted on benchmark datasets to highlight the defense and debiasing effect of the proposed De2GNN framework against adversarial attacks. The main contributions of this work are outlined as follows.
	
	\begin{enumerate}
		\item We empirically uncover a pervasive structural bias towards the prediction effect of nodes with low degree in current defense GNNs, prompting a comprehensive re-evaluation of the defense performance of these models.
		\item We propose a defense and debiasing framework, De2GNN, to mitigate the structural bias of nodes with low degree against graph adversarial attacks by incorporating the hetero-homo augmented graph and $k$NN augmented graph. Moreover, a node-wise attention mechanism has been employed to adaptively combine the information obtained from the hetero-homo augmented graph and $k$NN augmented graph.
		\item We conduct extensive experiments on three benchmark datasets to demonstrate the superiority of the proposed De2GNN, especially for the mitigation effect of the prediction bias on nodes with low degree. 
	\end{enumerate}
	
	The remainder of this work is summarized as follows. We review current efforts on the graph adversarial robustness including attacks and defenses on GNNs, and the mitigation of structural bias in Section \ref{sec:related}. The basic notations and definitions of GNNs, graph adversarial robustness, and structural bias are given in Section \ref{sec:pre}. An empirical study on the structural bias of current defense GNNs is conducted in Section \ref{sec:emp}. Based on the observations, we further proposed the defense and debiasing model, De2GNN, in Section \ref{sec:model}. Detailed experimental results and discussions are given in Section \ref{sec:exp}. Finally, we conclude this work in Section \ref{sec:con}.
	
	\section{Related Work}\label{sec:related}
	In this section, we briefly review the current efforts on graph adversarial robustness and structural bias mitigation on GNNs. We first provide an overview of the advancements in GNNs. Then, we present the graph adversarial robustness from both attack and defense perspectives. Finally, we introduce some representative works in bias mitigation in GNNs.

	\subsection{Graph Neural Networks}
	GNNs have succeeded greatly in recent years because of the neighborhood aggregation mechanism. The center node can obtain a more informative representation from the features of its connected neighbors than only utilizing its ego features through neighborhood aggregations. Based on this, several GNNs have been developed. For instance, Kipf {\em et al.} \cite{kipf2017semi} proposed graph convolutional network (GCN), one of the most classic graph neural networks, by utilizing a weighted aggregator based on the degree information of nodes. Hamilton {\em et al.} \cite{hamilton2017inductive} introduced GraphSAGE, which enabled inductive learning on large-scale graphs through neighborhood sampling and advanced aggregation operations, such as LSTM-based (Long Short-Term Memory) aggregator to enhance the scalability of GNNs. Recognizing that different neighbors in a neighborhood may impact downstream tasks differently, Veli{\v{c}}kovi{\'c} {\em et al.} \cite{velivckovic2017graph} further proposed graph attention network (GAT), which adaptively decided the aggregated weights of each neighbor via the self-attention mechanism. Wu {\em et al.} \cite{wu2019simplifying} presented simplified graph convolutional networks (SGC) by removing the non-linear activation functions in the middle layers of GCN, which achieved a comparable prediction performance but required less time and space complexity. Besides the above classic GNNs, various different GNNs have been proposed to enhance their abilities on convolutional depth \cite{liu2020towards1,li2020deepergcn}, scalability \cite{serafini2021scalable,md2021distgnn,wan2023scalable}, and other aspects.

	\subsection{Graph Adversarial Robustness}
	
	Despite their remarkable success, GNNs are vulnerable to malicious adversarial attacks. Specifically, graph adversarial attacks refer to the scenarios in which attackers try to inject some unnoticeable perturbations, such as adding or removing links, modifying node features, etc., to the original clean graph, aiming to mislead the prediction of GNNs in corresponding downstream tasks. Z\"ugner {\em et al.} \cite{zugner2018adversarial} proposed the first graph attack strategy, Nettack, by greedily selecting the adversarial links based on the reduction of prediction margins between the ground truth label and max-possible wrong label. After that, Dai {\em et al.} \cite{dai2018adversarial} introduced GraphArgmax using reinforcement learning algorithms. Z\"ugner {\em et al.} \cite{zugner2018adversarial1} further proposed Metattack by utilizing the meta gradients of meta-learning to achieve global attacks on the whole graph. Besides directly modifying the original graph, recent studies also investigated the possibilities of injecting some fake nodes, aiming to reach more practical and realistic attacks in real-world scenarios. Li {\em et al.} \cite{li2020adversarial} proposed SGA to achieve the adversarial attacks on large-scale graphs by focusing on $k$-hop subgraphs, vastly reducing the space complexity. Sun {\em et al.} \cite{sun2019node} presented NIPA by modeling the neighborhood selections of the newly generated nodes as a Markov process and then employed a reinforcement learning algorithm to achieve the fake node injections. Fang {\em et al.} \cite{fang2024gani} introduced GANI, which utilized a more reasonable feature generation mechanism and evolutionary optimization-based neighborhood selection to achieve more imperceptible node injection attacks than previous methods.

	To eliminate the negative effect of adversarial attacks, researchers also put lots of effort into defending the possible attacks. Specifically, current defense methods can be broadly divided into three categories, namely, pre-processing methods, robust aggregators, and adversarial training. From the view of pre-processing, defenders try to detect possible perturbations generated by attackers. Wu {\em et al.} \cite{wu2019adversarial} proposed a simple yet effective defense method by removing the corresponding links where the connected neighbors have low Jaccard feature similarity. Moreover, by empirically observing that the perturbations injected by attackers tend to be high-rank attacks, Entezari {\em et al.} \cite{entezari2020all} introduced an SVD model that obtained a low-rank appropriate adjacency matrix via singular value decomposition. From the view of robust aggregators, defenders aim to design a more robust aggregator to automatically give lower weights to the perturbed information. Zhu {\em et al.} \cite{zhu2019robust} modeled the representations of nodes from the aggregations of GNNs as Gaussian distributions and then gave lower aggregation weights to nodes with higher variance, assuming that these corresponding links tend to be injected by attackers. Chen {\em et al.} \cite{ijcai2021p310} proposed a Median model by only aggregating the median parts of neighborhood features per dimension. From the view of adversarial training, which aims to improve the tolerance of models by actively employing some possible perturbations in the training phase. Deng {\em et al.} \cite{deng2023batch} proposed the batch virtual adversarial training to smooth the outputs of GNNs, while Jin {\em et al.} \cite{jin2021robust} applied adversarial training on the latent representations of nodes to improve the robustness and generalization ability of GNNs. Xu {\em et al.} \cite{xu2019topology} introduced a structure-based adversarial training by generating the adversarial links via topology attack methods.
	
	However, the same structural bias of current GNNs on the clean graph, which has a worse prediction performance on nodes with lower degree, has also been empirically observed in current defense models. Hence, although we are able to obtain a better prediction performance globally by utilizing current defense methods, the prediction performance on nodes with low degree is also still not so promising. Therefore, in this work, we aim to mitigate the structural bias that exists in current defense models against adversarial attacks.
	
	\subsection{Structural Bias Mitigation in Graph Neural Networks}
	
	Besides the poor robustness, recent studies also pointed out that current GNNs still confront the structural bias on nodes with low degree. Specifically, the structural bias in current GNNs indicates the prediction performance on nodes with different levels of degree. GNNs tend to have a much better prediction performance on nodes with high degree than on nodes with low degree. The above situation is intuitively reasonable as the performance of GNNs largely depends on the effect of neighborhood aggregations. It is clear that nodes with low degree have significantly less neighborhood information than nodes with high degree.
	
	To tackle the structural bias, i.e., degree bias, in current GNNs, a series of improved GNNs have been proposed. For instance, Tang {\em et al.} \cite{tang2020investigating} investigated the degree bias of GNNs and further proposed a mitigation strategy by inducing learnable parameters to encode the degree information of nodes during the neighborhood aggregations. Moreover, an additional recurrent neural network was employed to learn the relationships between different degree values. Liu {\em et al.} \cite{liu2021tail} proposed Tail-GNN by utilizing the information from nodes with high degree as a supervised signal to guide nodes with low degree to learn the missing information. Yun {\em et al.} \cite{yun2022lte4g} introduced LTE4G by first assigning an expert teacher GNN to each balanced subset of nodes considering both the class and degree perspectives, then adopting knowledge distillation to obtain the corresponding student GNNs for classifying nodes in the corresponding subsets. Moreover, Kang {\em et al.} \cite{kang2022rawlsgcn} utilized the theory of distributive justice to balance the utility of nodes with high and low degrees in the loss function, and then proposed the pre-processing and in-processing methods based on the gradients of weighted matrices. Recently, Liu {\em et al.} \cite{liu2023generalized} introduced a novel method by removing the redundant information of nodes with high degree while enriching the missing information of nodes with low degree to achieve a degree fair GNN. However,  the above studies rarely considered that the graph data may be noisy due to adversarial attacks. Directly employing the above methods to a perturbed graph may not achieve a satisfactory debiasing performance, as they are all designed without considering the possible adversarial attacks. Therefore, in this work, we aim to mitigate the prediction performance bias existing in the nodes with low degree in the adversarial attack scenario.

	\section{Preliminaries}\label{sec:pre}
	In this section, we introduce the basic definitions of graph neural networks and graph adversarial robustness.

	\subsection{Graph Neural Networks}
	A graph can be modeled as $G = (V, E, X)$, where $V$, $E$, and $X$ represent the node set, link set, and feature matrix, respectively. In particular, we denote the number of nodes as $|V|$, and the number of links as $|E|$. Moreover, we take the node classification as the representative downstream task in this work. To tackle node classification problems, a GNN model $f_{\theta^*}$ can be obtained based on training nodes $V_{\rm train} \subset V $ by minimizing the loss function $\mathcal{L}_{\rm train}$. Specifically, we denote a general $k$-layer GNN as
	
	\begin{equation}\label{eq:gcn}
		h_u^{(k)} = \sigma(W^{(k)} \cdot Agg(h_v^{(k-1)} | v \in \mathcal{N}(u)),
	\end{equation}
	where $h_u^{(k)}$ indicates the representation of node $u$ in the $k$-th layer, and $\mathcal{N}(u)$ denotes the neighbors of node $u$ with a self-loop. $\sigma$ is the activation function like ReLU. $W^{(k)}$ is the parameter to be optimized in the $k$-th layer. Particularly, $Agg(\cdot)$ is the aggregation function that aggregates the information of neighboring nodes. Taking the aggregation mechanism of GCN \cite{kipf2017semi} which utilizes weighted aggregations of degree information of nodes as an example, we can have

	\begin{equation}\label{eq:agg}
		Agg(h_v^{(k-1)} | v \in \mathcal{N}(u)) = \sum_{v \in \mathcal{N}(u)} \frac{1}{\sqrt{|\mathcal{N}(u)| \cdot |\mathcal{N}(v)|}} \cdot h_v^{(k-1)},
	\end{equation}
	where $|\mathcal{N}(u)|$ ($|\mathcal{N}(v)|$) represents the number of neighbors of node $u$ ($v$).
	
	After we obtain the hidden representation in the last aggregation layer, we can utilize the classic cross-entropy loss to obtain the optimal model parameters $\theta$, i.e.,
	\begin{equation}\label{eq:loss_train}
		\min \limits_{\theta} \mathcal{L}_{\rm train} = - \sum_{v \in V_{\rm train}} \ln Z_{v,t}, \quad Z = f_\theta(G),
	\end{equation}
	where $Z_{u,t}$ is the predicted probability of node $u$ belonging to its true label $t$.
	
	\subsection{Graph Adversarial Robustness}
	
	Graph adversarial robustness can be broadly divided into graph adversarial attacks and defenses. The former aims to deceive a GNN model by injecting small perturbations, such as link and feature noises, into the original graph $G$. The latter pays attention to improving the robustness of the specific GNN model against such perturbations injected by attackers.
	
	From the perspective of attacks, we can formulate the objectives of adversarial attacks as
	\begin{equation}\label{eq:loss_attack}
		\begin{aligned}
			& 	\min \limits_{\alpha} \mathcal{L}_{\rm atk}(f_{\theta}(G^{\prime}))
			\\
			& s.t. G^{\prime} = {Att}_{\alpha} (G) \quad {\rm and} \quad |G^{\prime} - G| \leq \Delta,
		\end{aligned}
	\end{equation}
	where $f_{\theta}$ represents the (surrogate) GNN model being attacked. We can further divide graph adversarial attacks into evasion attacks and poisoning attacks by deciding whether to retrain the GNN model after attacks or not. $Att(\cdot)$ denotes the specific attack model. $\mathcal{L}_{\rm atk}$ represents the attack loss of the attackers and $\alpha$ denotes the optimal parameters of the attack model. Particularly, in this work, we focus on global attacks that aim to influence the overall classification performance of all nodes. Taking the Metattack strategy \cite{zugner2018adversarial1} focusing on global poisoning attacks as an example, the overall loss function of the attacks can be given as 
	\begin{equation}\label{eq:loss_attack1}
		\mathcal{L}_{\rm atk} = - \mathcal{L}_{\rm train} = \sum_{v \in V_{\rm train}} \ln Z_{v, t} \quad Z = f_{\theta}(G^{\prime}).
	\end{equation}
	Metattack aims to minimize $\mathcal{L}_{\rm atk}$ using a surrogate GNN model $f_{\theta_1}$, i.e., SGC model \cite{wu2019simplifying}, by treating the graph structure matrix $A$ as a hyperparameter based on the guidance of meta gradients. As a result, Metattack tends to decrease the probabilities of the corresponding nodes belonging to their ground truth label. 
	
	On the other side, the purpose of graph adversarial defenses is to reduce the negative effect of adversarial attacks as much as possible. As discussed in Section \ref{sec:related}, current defense models can be broadly divided into pre-processing methods, robust aggregators, and adversarial training. By taking pre-processing operation as an example, the defenses of GNNs are similar to (\ref{eq:loss_train}), i.e.,
	
	\begin{equation}\label{eq:loss_def}
		\min \limits_{\beta} \mathcal{L}_{\rm def} = - \sum_{v \in V_{\rm train}} \ln Z_{v,t}, \quad Z = f_{\theta}({\rm Def}_{\beta}(G^{\prime})),
	\end{equation}
	where ${\rm Def}(\cdot)$ refers to the corresponding defense model which tries to detect and then removes the possible noisy structures in graph $G^{\prime}$ under the guidance of the optimal parameter $\beta$.

	\section{Structural Bias in Graph Adversarial Defense}\label{sec:emp}

	\begin{figure}[t]
		\centering
		\subfigure{
			\begin{minipage}[]{0.5\linewidth}
				\includegraphics[scale=0.25]{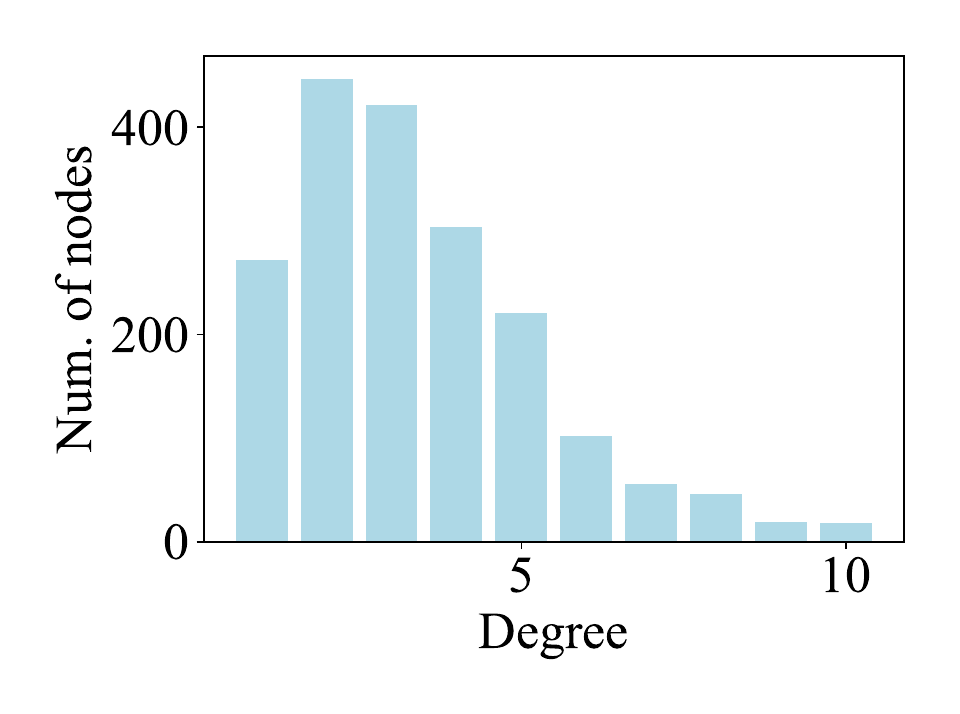}
			\end{minipage}%
		}%
		\subfigure{
			\begin{minipage}[]{0.5\linewidth}
				\includegraphics[scale=0.25]{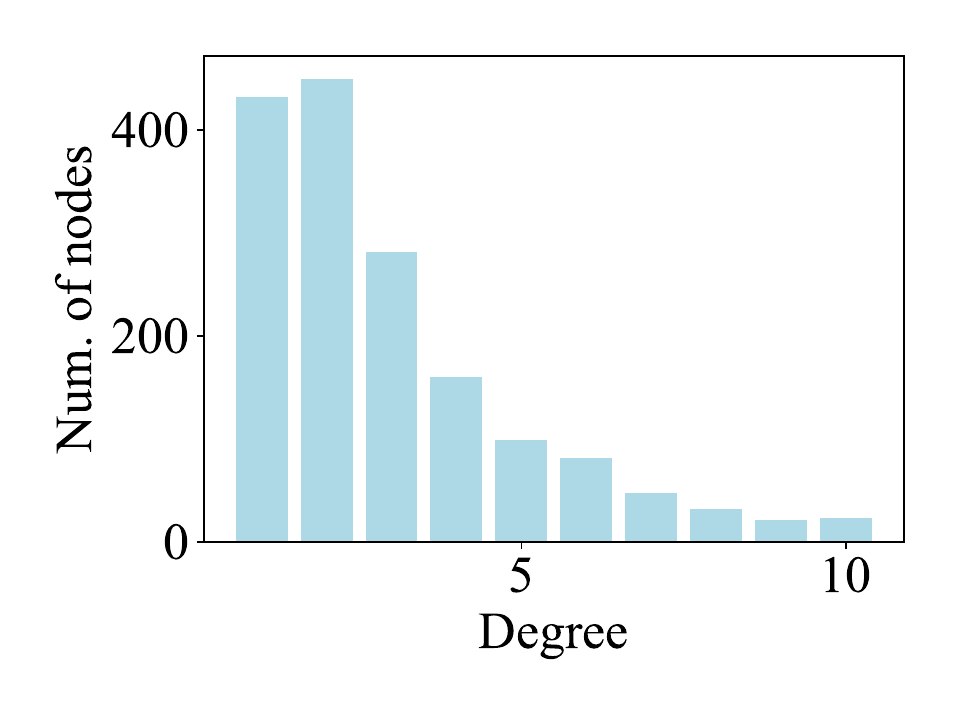}
			\end{minipage}%
		}%
		\caption{Degree distributions of Cora and Citeseer datasets.}
		\label{fig:degree_freq}
	\end{figure}

	\begin{figure}[t]
		\centering
		\subfigure{
			\begin{minipage}[]{0.5\linewidth}
				\includegraphics[scale=0.25]{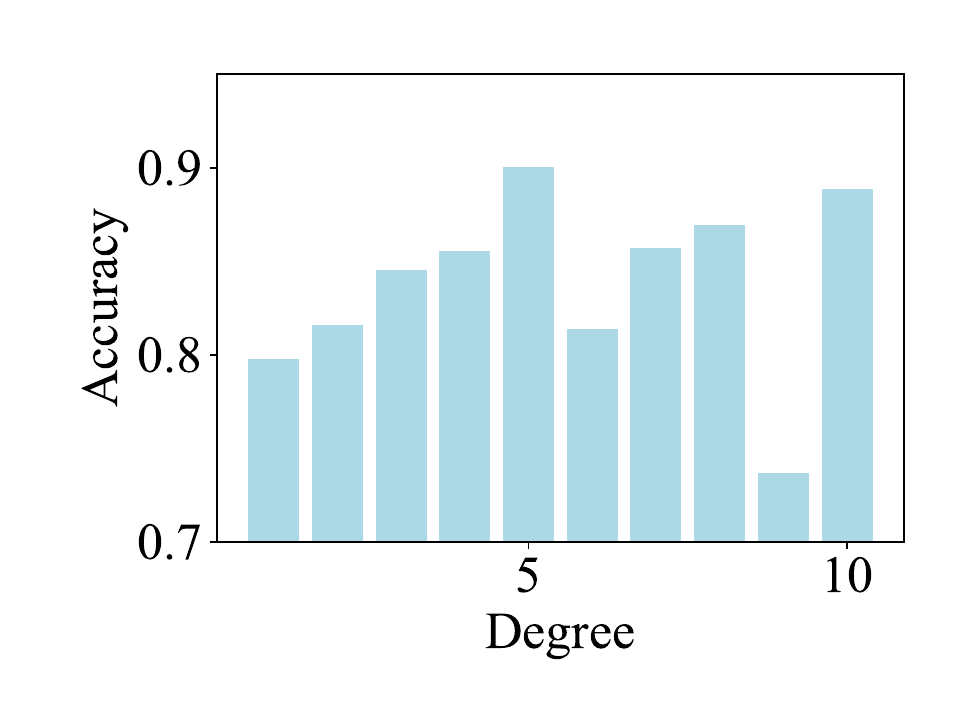}
			\end{minipage}%
		}%
		\subfigure{
			\begin{minipage}[]{0.5\linewidth}
				\includegraphics[scale=0.25]{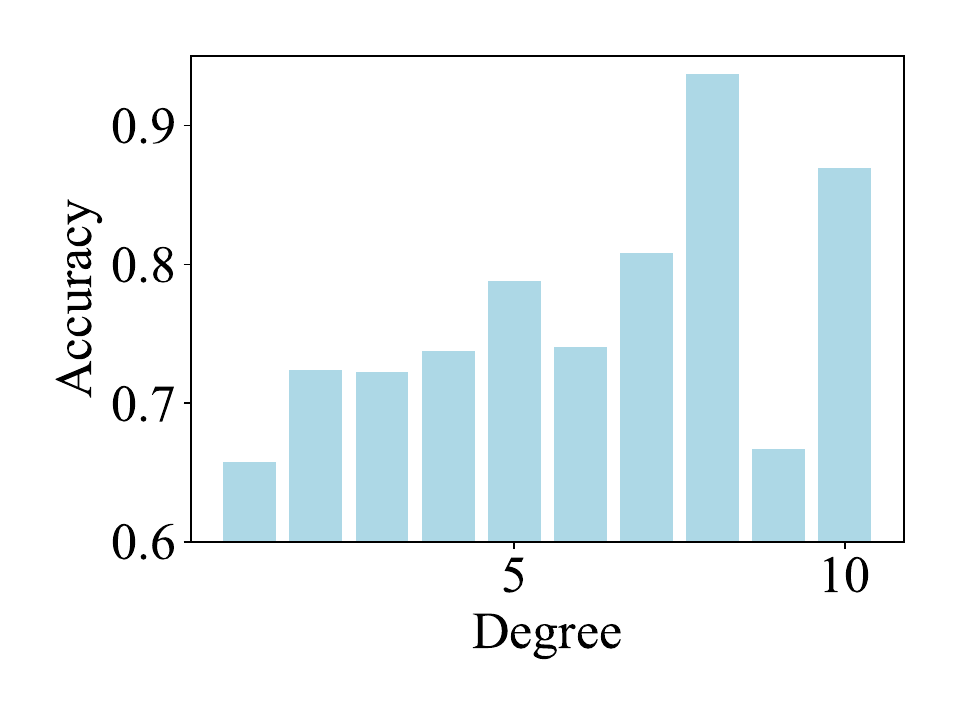}
			\end{minipage}%
		}%
		\caption{Accuracy of GCN with regard to node degree on Cora and Citeseer datasets without attacks.}
		\label{fig:acc_clean}
	\end{figure}

	\begin{figure}[t]
		\centering
		\subfigure{
			\begin{minipage}[]{0.5\linewidth}
				\includegraphics[scale=0.25]{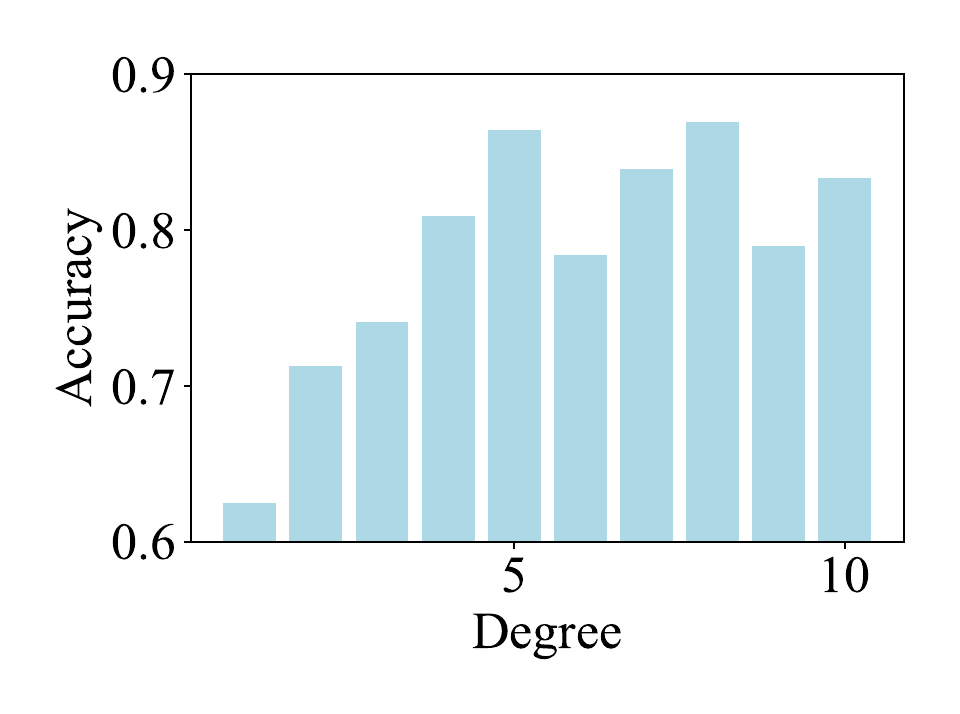}
			\end{minipage}%
		}%
		\subfigure{
			\begin{minipage}[]{0.5\linewidth}
				\includegraphics[scale=0.25]{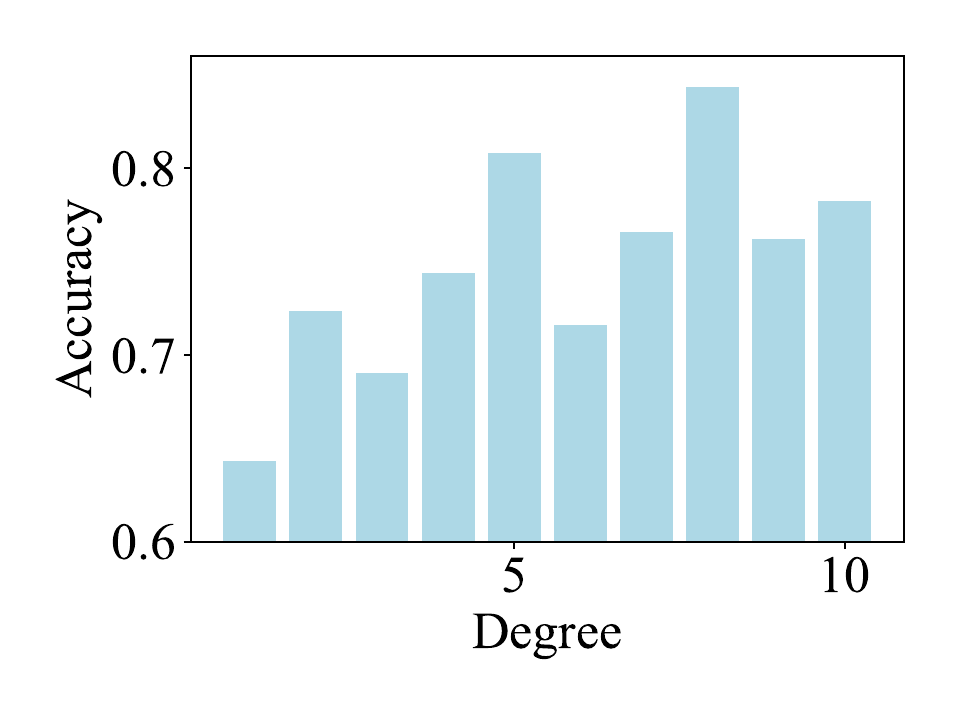}
			\end{minipage}%
		}%
		\caption{Accuracy of Jaccard with regard to node degree on Cora and Citeseer datasets by injecting 25\% of noisy links via Metattack.}
		\label{fig:acc_attack_jaccard}
	\end{figure}

	\begin{figure}[t]
		\centering
		\subfigure{
			\begin{minipage}[]{0.5\linewidth}
				\includegraphics[scale=0.25]{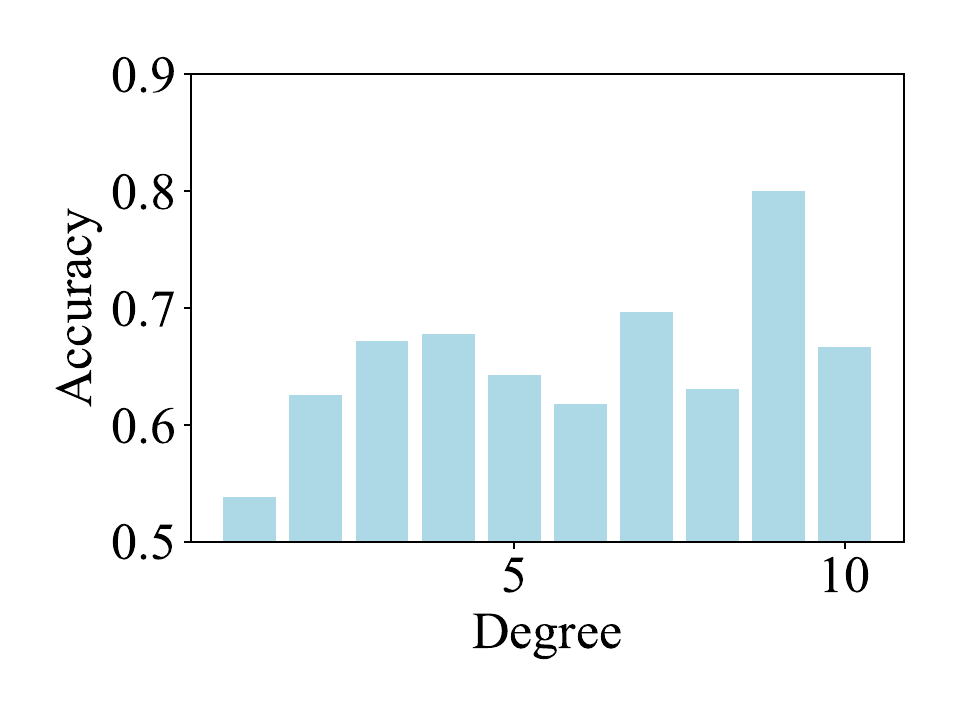}
			\end{minipage}%
		}%
		\subfigure{
			\begin{minipage}[]{0.5\linewidth}
				\includegraphics[scale=0.25]{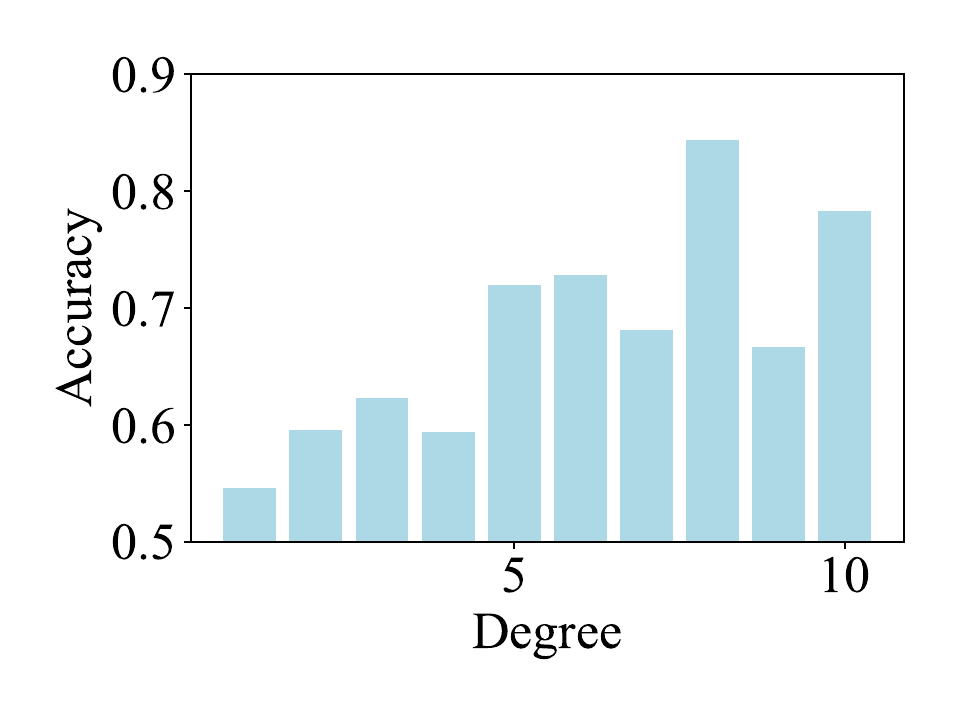}
			\end{minipage}%
		}%
		\caption{Accuracy of SVD with regard to node degree on Cora and Citeseer datasets by injecting 25\% of noisy links via Metattack.}
		\label{fig:acc_attack_svd}
	\end{figure}

	Structural bias, e.g., degree bias considered in this work, in the predictions of GNNs refers to the issue that current GNNs tend to have a prediction bias between nodes with high degree and low degree. Specifically, by giving a degree threshold, we can broadly divide the nodes in a graph $G$ into two groups, i.e., high degree group $HD$ and low degree group $LD$ based on the degree of each node by a threshold degree value. The structural bias is that HD usually obtains a higher prediction accuracy than LD, i.e., $ACC(HD) \gg ACC(LD)$. Therefore, our goal is to improve the prediction accuracy of nodes with low degree (i.e., $ACC(LD)$), especially in the adversarial defense scenario.

	To mitigate such bias in current defense models, we first demonstrate the long tail effect of typical networks. Fig. \ref{fig:degree_freq} shows the node frequency with regard to node degree on Cora and Citeseer datasets. The statistics of Cora and Citeseer datasets will be given in Section \ref{sec:exp}. We can observe that most of the nodes have low degree in both datasets, demonstrating that the long tail effect truly exists in the graphs.  Besides the relationship between the node frequency and degree, we also present the prediction performance of GCN with regard to node degree, as shown in Fig. \ref{fig:acc_clean}. We can observe that nodes with lower degree, which constitute most of the network, tend to have poorer classification accuracy than nodes with higher degree on both datasets. The above results are reasonable as the core design of GNNs is the neighborhood aggregation. Due to the sparse neighborhood of nodes with low degree, they cannot obtain enough useful features for characterizing themselves, thus resulting in poor prediction performance.
	
	More importantly, we further conduct a similar experiment on several defense GNNs under the perturbed datasets by injecting 25\% of noisy links generated from the Metattack method \cite{zugner2018adversarial1}. Figs. \ref{fig:acc_attack_jaccard} and \ref{fig:acc_attack_svd} show the prediction results of two defense models, i.e., Jaccard \cite{wu2019adversarial} and SVD \cite{entezari2020all},  on the perturbed datasets, respectively. We can also observe a similar or even severe prediction bias on nodes with low degree, indicating that current defense models also suffer the same structural bias problem. Therefore, in the next section, we propose a defense and debiasing GNN, De2GNN, aiming to achieve better debiasing performance under adversarial scenarios.

	\begin{figure*}[t]
		\centering
		\includegraphics[width=1\textwidth]{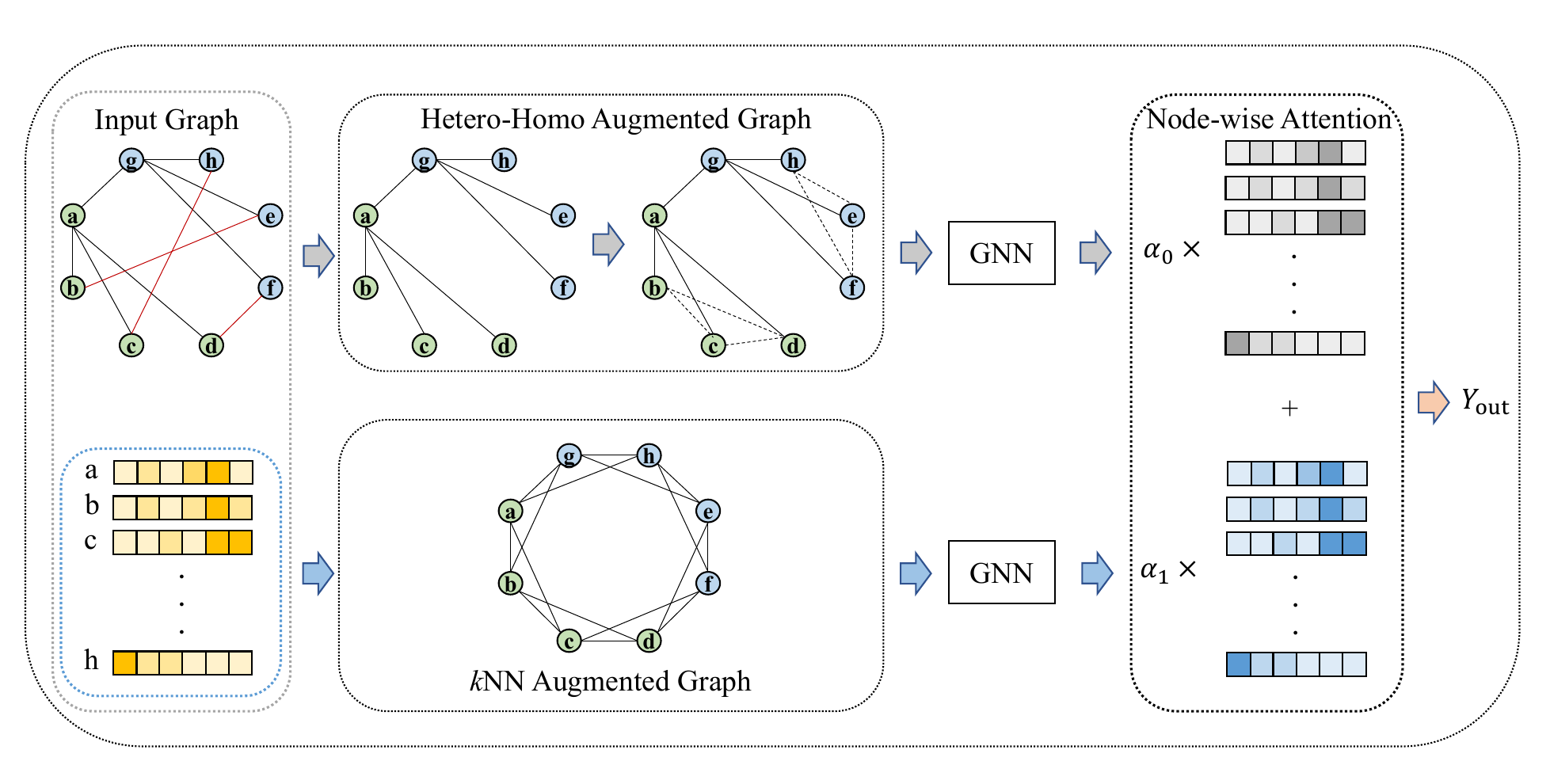} 
		\caption{Systematic framework of De2GNN. Red links in the input graph indicate corresponding adversarial noises.}
		\label{fig:framework}
	\end{figure*}
	
	\section{Proposed Model}\label{sec:model}
	Based on the observations from Section \ref{sec:emp}, we know that current defense methods also suffer the structural prediction bias on nodes with low degree. To address this issue, we propose a simple yet effective graph neural network, De2GNN, to {\em de}fend off adversarial attacks and {\em de}bias the structural bias at the same time. Specifically, the proposed De2GNN includes three modules, including the hetero-homo augmented graph construction, $k$NN augmented graph construction, and multi-view node-wise attention mechanism. The details of these modules are given as follows. 
	
	\subsection{Hetero-Homo Augmented Graph Construction}
	\subsubsection{Removal of Heterophilic Links}
	By giving a possibly perturbed graph $G^\prime$, we first aim  to detect and remove potential adversarial links. Based on previous studies \cite{bojchevski2019certifiable, jin2020adversarial}, attackers usually add some heterophilic links where connected nodes tend to share dissimilar features. Thus, similar to the idea of the Jaccard model \cite{wu2019adversarial}, some similarity metrics are used to detect potential adversarial links according to the feature similarity of connected nodes. In particular, as attackers usually tend to add new links to graphs, we only focus on the adversarial link removal operation in this step. It is worth noting that the proposed De2GNN is also robust to remove-link attacks to some extent as we will add homophilic links in the next step. 
	
	Specifically, we select the Jaccard similarity and Cosine similarity as feature similarity evaluation metrics for the graph with binary and continuous features, respectively. The similarity of a given link $E_{ij}$, $s_{ij}$, is represented as
	\begin{equation}
		s_{ij} = \begin{cases}
			\frac{|X_i \cap X_j|}{|X_i \cup X_j|}, & \text{if $X$ is discrete} \\ \\
			\frac{X_i \cdot X_j}{\|X_i\| \cdot \|X_j\|}, & \text{if $X$ is continuous}
		\end{cases},
	\end{equation}
	where $X_i$ ($X_j$) denotes the feature vector of node $i$ ($j$).
	
	The link which has a feature similarity lower than the predefined threshold $t_1$ will be considered as a heterophilic link and be further removed from $G^\prime$. Finally, we obtain a new graph $G_{-{\rm hetero}}=(V,E_{-{\rm hetero}},X)$, where the latest link set $E_{-{\rm hetero}}$ is given by 
	\begin{equation}
		E_{-{\rm hetero}} = \{(i, j) \in E | s_{ij} > t_1\}.
	\end{equation}

	\subsubsection{Addition of Homophilic Links for Tail Nodes}
	Through removing heterophilic links in the previous step, we can obtain a relatively cleaner graph $G_{-{\rm hetero}}$. However, as we only focus on the link removal operation while the long tail effects have already existed in the original graph, a large part of nodes tend to have low degree. Therefore, in this step, we try to enrich the neighborhood of nodes with low degree. We assume that the original link set in this step is $E_{+{\rm homo}} = E_{-{\rm hetero}}$. Specifically, we first train a surrogate GNN, $M_{{\rm sur}}$, based on graph $G_{-{\rm hetero}}$ as it has removed the potential noise to some extent. Then, for each tail node $v$ whose prediction confidence is higher than a predefined threshold $t_2$, we greedily connect it with top $p$ nodes which have the highest probabilities that belong to the same class as $v$. By doing this, we can improve the homophily level of original $G_{-{\rm hetero}}$, which is beneficial to the neighborhood aggregations for GNN models. The new potential link set for the specific tail node $v$ is denoted as
	\begin{equation}
		E_{+{\rm homo}}(v) = \{(v, q) | q \in N_{\rm top}(M_{{\rm sur}}, v, p)\},
	\end{equation}
	where $N_{\rm top}(M_{{\rm sur}}, v, p)$ represents the top $p$ nodes having the highest prediction confidence that belong to the same class as node $v$ based on the model $M_{\rm sur}$. Then, these newly added potential links are combined with original links to construct the augmented link set, i.e.,
	\begin{equation}
		E_{+{\rm homo}} = E_{+{\rm homo}} \cup E_{+{\rm homo}}(v).
	\end{equation}
	After this, we can obtain a relatively cleaner yet more informative graph $G_{+{\rm homo}}=(V, E_{+{\rm homo}}, X)$ which has removed potential adversarial links while enriching the sparse neighborhoods for tail nodes.

	\subsection{kNN Augmented Graph Construction}
	Although we have removed the potential noise in the refined graph $G_{+{\rm homo}}$, there may still exist some concealed noise. Thus, we further utilize raw features of nodes of the graph to obtain an attack-agnostic augmented graph $G_{k{\rm NN}}$ based on the classic $k$ nearest neighbors (i.e., $k$NN) algorithm \cite{peterson2009k}. In the graph $G_{k{\rm NN}}$, each node will be connected to $k$ nodes that have the highest feature similarity.
	
	\subsection{Multi-view Node-wise Attention}
	Based on the above two modules, we obtain two graphs (i.e., $G_{+{\rm homo}}$ and $G_{k{\rm NN}}$) from different augmentation views. This step combines these two graph views to learn the final representation. Specifically, De2GNN first employs two separate GNNs to learn independent node representations in these two graph views, respectively. Then, De2GNN adopts an attention mechanism from a node-wise level to adaptively determine weights of representations from different views for each node. The details can be formulated as 
	\begin{equation}
		H_{+{\rm homo}} = {\rm GNN_1}(G_{+homo}),
	\end{equation}
	\begin{equation}
		H_{k{\rm NN}} = {\rm GNN_2}(G_{k{\rm NN}}),
	\end{equation}	
	\begin{equation}
		\hat{\alpha} = \sigma([H_{+{\rm homo}}, H_{k{\rm NN}}] \cdot W),
	\end{equation}	
	\begin{equation}
		\alpha = {\rm softmax}(\hat{\alpha}),
	\end{equation}
	where $[\cdot, \cdot]$ denotes the concatenation operation. $W$ is the learnable weighted matrix. Moreover, attention scores are normalized via the softmax function. Then, the final representation can be obtained as 
	\begin{equation}
		Y^{\rm final} = \alpha_{\cdot, 0} \cdot H_{+{\rm homo}} + \alpha_{\cdot, 1} \cdot H_{k{\rm NN}},
	\end{equation}	
	where $\alpha_{\cdot, 0}$ and $\alpha_{\cdot, 1}$ represent attention scores for the information from the hetero-homo augmented graph and $k$NN augmented graph, respectively. $Y^{\rm final}$ denotes the final outputs of De2GNN.

	\subsection{Overview}
	The overall framework of the proposed De2GNN is shown in Fig. \ref{fig:framework}. The original (possibly perturbed) graph is first employed to remove heterophilic links based on feature similarities and to add homophilic links for tail nodes based on the prediction confidence from a surrogate GNN. At the same time, an augmentation graph generated based on potential neighbors obtained from the $k$NN algorithm has been constructed. Then, these two graphs are fed into two separate GNNs to learn independent node representations, respectively. Finally, a node-wise attention mechanism is adopted to adaptively determine the weight of information from different views for each node, thereby obtaining the final outputs.

	\section{Experiments}\label{sec:exp}
	In this section, we conduct comprehensive experiments to evaluate the defense and debiasing effectiveness of the proposed De2GNN. Specifically, we first introduce the benchmark datasets, baseline methods, and experimental setups. Then, we present the experimental results and corresponding discussions.
	
	\subsection{Datasets}
	We evaluated the proposed strategies on three representative benchmark datasets including Cora, Citeseer, and Pubmed \cite{sen2008collective}. Following the settings in the previous study \cite{dai2018adversarial}, we extracted the largest connected component of them, and the statistical information is given in Table \ref{tab:statistic}.

	\begin{table}[]
		\centering
		\caption{Statistics of datasets.}
		\resizebox{\linewidth}{!}{
			\begin{tabular}{c|c|c|c|c|c}
				\bottomrule
				\textbf{\rule{0in}{0.15in}Datasets} &
				\textbf{\#Nodes} &
				\textbf{\#Links} &
				\textbf{\#Features} &
				\textbf{\#Classes} &
				\textbf{Avg. Degree} \\ \hline\hline
				\rule{0in}{0.15in}Cora     & 2,485  & 5,069  & 1,433 & 7 & 4.08 \\ 
				Citeseer & 2,100  & 3,668  & 3,703 & 6 & 3.48 \\
				Pubmed   & 19,717 & 44,324 & 500   & 3 & 4.50 \\ \hline
		\end{tabular}}%
		
		\label{tab:statistic}
		
	\end{table}

	\begin{table*}[tbp]
		\centering
		\caption{Accuracy of all nodes and tail nodes (i.e., degree $\leq$ 5) of baselines on clean datasets without attacks. The best results in each row are boldfaced.}
		\label{tab:acc_before}
		\resizebox{0.9\linewidth}{!}{\begin{tabular}{c|c|c|c|c|c|c|c}
				\bottomrule
				\textbf{\rule{0in}{0.15in}Datasets} & \textbf{Types} & \textbf{GCN} & \textbf{Jaccard} & \textbf{SVD} & \textbf{RGCN} & \textbf{Median} & \textbf{De2GNN} \\
				\hline\hline
				\multirow{2}{*}{Cora} & \rule{0in}{0.15in}ALL & 0.8341\scriptsize{$\pm$0.0031}  & \textbf{0.8458\scriptsize{$\pm$0.0038}} & 0.7403\scriptsize{$\pm$0.0050}  & 0.8477\scriptsize{$\pm$0.0041} & 0.8405\scriptsize{$\pm$0.0027} & 0.8242\scriptsize{$\pm$0.0067} \\
				\cline{2-8}
				& Tail & 0.8319\scriptsize{$\pm$0.0039}  & 0.8429\scriptsize{$\pm$0.0038} & 0.7408\scriptsize{$\pm$0.0070}  & 0.8380\scriptsize{$\pm$0.0034} & \textbf{0.8441\scriptsize{$\pm$0.0040}} & 0.8241\scriptsize{$\pm$0.0063} \\
				\hline
				\multirow{2}{*}{Citeseer} & ALL & 0.7129\scriptsize{$\pm$0.0029}  & 0.7134\scriptsize{$\pm$0.0078}  & 0.6994\scriptsize{$\pm$0.0076}  & 0.7214\scriptsize{$\pm$0.0041} & 0.7220\scriptsize{$\pm$0.0063} & \textbf{0.7270\scriptsize{$\pm$0.0028}} \\
				\cline{2-8}
				& Tail & 0.7014\scriptsize{$\pm$0.0038}  & 0.7015\scriptsize{$\pm$0.0083}  & 0.6835\scriptsize{$\pm$0.0084}  & 0.7088\scriptsize{$\pm$0.0048} & 0.7128\scriptsize{$\pm$0.0055} & \textbf{0.7156\scriptsize{$\pm$0.0029}} \\
				\hline
				\multirow{2}{*}{Pubmed} & ALL & \textbf{0.8603\scriptsize{$\pm$0.0021}}  & \textbf{0.8603\scriptsize{$\pm$0.0020}}  & 0.8404\scriptsize{$\pm$0.0024}  & 0.8524\scriptsize{$\pm$0.0011} &  0.8437\scriptsize{$\pm$0.0015} & 0.8558\scriptsize{$\pm$0.0020}\\
				\cline{2-8}
				& Tail & \textbf{0.8578\scriptsize{$\pm$0.0026}}  & 0.8577\scriptsize{$\pm$0.0025}  & 0.8431\scriptsize{$\pm$0.0030}  & 0.8495\scriptsize{$\pm$0.0011} &  0.8369\scriptsize{$\pm$0.0017} & 0.8568\scriptsize{$\pm$0.0020}\\
				\toprule
		\end{tabular}}
	\end{table*}

	\begin{table*}[htbp]
		\centering
		\caption{Accuracy of all nodes and tail nodes (i.e., degree $\leq$ 5) of baselines on perturbed datasets by injecting 25\% of noisy links via Metattack. The best results in each row are boldfaced.}
		\label{tab:acc_after}
		\resizebox{0.9\linewidth}{!}{\begin{tabular}{c|c|c|c|c|c|c|c}
				\bottomrule
				\textbf{\rule{0in}{0.15in}Datasets} & \textbf{Types} & \textbf{GCN} & \textbf{Jaccard} & \textbf{SVD} & \textbf{RGCN} & \textbf{Median} & \textbf{De2GNN} \\
				\hline\hline
				\multirow{2}{*}{Cora} & \rule{0in}{0.15in}ALL & 0.4570\scriptsize{$\pm$0.0074}  & 0.7231\scriptsize{$\pm$0.0076} & 0.6447\scriptsize{$\pm$0.0078}  & 0.5816\scriptsize{$\pm$0.0128} & 0.5874\scriptsize{$\pm$0.0264} & \textbf{0.7538\scriptsize{$\pm$0.0097}} \\
				\cline{2-8}
				& Tail & 0.4396\scriptsize{$\pm$0.0077}  & 0.7135\scriptsize{$\pm$0.0087} & 0.6408\scriptsize{$\pm$0.0082}  & 0.5681\scriptsize{$\pm$0.0135} & 0.5818\scriptsize{$\pm$0.0263} & \textbf{0.7491\scriptsize{$\pm$0.0101}} \\
				\hline
				\multirow{2}{*}{Citeseer} & ALL & 0.4992\scriptsize{$\pm$0.0066}  & 0.7024\scriptsize{$\pm$0.0087}  & 0.6271\scriptsize{$\pm$0.0110}  & 0.5809\scriptsize{$\pm$0.0038} & 0.5835\scriptsize{$\pm$0.0087} & \textbf{0.7171\scriptsize{$\pm$0.0031}} \\
				\cline{2-8}
				& Tail & 0.4777\scriptsize{$\pm$0.0051}  & 0.6929\scriptsize{$\pm$0.0101}  & 0.6024\scriptsize{$\pm$0.0120}  & 0.5638\scriptsize{$\pm$0.0032} & 0.5657\scriptsize{$\pm$0.0088} & \textbf{0.7082\scriptsize{$\pm$0.0036}} \\
				\hline
				\multirow{2}{*}{Pubmed} & ALL & 0.5650\scriptsize{$\pm$0.0034}  & 0.8346\scriptsize{$\pm$0.0018}  & 0.8382\scriptsize{$\pm$0.0010}  & 0.4503\scriptsize{$\pm$0.0051} &  0.5080\scriptsize{$\pm$0.0773} & \textbf{0.8470\scriptsize{$\pm$0.0039}}\\
				\cline{2-8}
				& Tail & 0.5794\scriptsize{$\pm$0.0044}  & 0.8284\scriptsize{$\pm$0.0023}  & 0.8399\scriptsize{$\pm$0.0011}  & 0.4392\scriptsize{$\pm$0.0055} &  0.5063\scriptsize{$\pm$0.0746} & \textbf{0.8487\scriptsize{$\pm$0.0035}}\\
				\toprule
		\end{tabular}}
	\end{table*}

	\subsection{Baselines}
	To comprehensively evaluate the defense and debiasing effect of De2GNN, we select several classic and defense GNNs, including GCN, Jaccard, SVD, RGCN, and Median as baselines.
	\begin{enumerate}
		\item \textbf{GCN} \cite{kipf2017semi}: GCN is a traditional GNN that obtains the low dimensional representation of nodes via aggregating the local structural and feature information of neighbors.

		\item \textbf{Jaccard} \cite{wu2019adversarial}: Jaccard is a defense GNN by employing a pre-process operation on the basic GCN. Specifically, based on the assumption that connected node pairs with low similarity tend to be noise, Jaccard first calculates the Jaccard similarity of connected node pairs and only preserves those links with higher similarity.
		
		\item \textbf{SVD} \cite{entezari2020all}: SVD is a defense strategy based on low-rank graph approximation centered around Singular Value Decomposition, effectively mitigating the impact of adversarial attacks and enhancing the robustness of GNN models.

		\item \textbf{RGCN} \cite{zhu2019robust}: RGCN utilizes Gaussian distributions as hidden node representations instead of traditional vectors. This adaptation enables RGCN to automatically absorb the effects of adversarial changes on the variances of the Gaussian distributions. Additionally, RGCN introduces a variance-based attention mechanism to mitigate the propagation of adversarial attacks within GCNs, assigning lower aggregation weights to nodes with higher variance, thereby addressing the injection of adversarial links by attackers.

		\item \textbf{Median} \cite{ijcai2021p310}: Median proposes to aggregate only the median part of features from the neighborhood per dimension, aiming to mitigate the impact of adversarial perturbations based on the breaking point theory.
	\end{enumerate}
	
	\subsection{Setups} 
	Following the previous work \cite{zugner2018adversarial}, each dataset is randomly split into the training set (10\%), validation set (10\%), and test set (80\%). In the following experiments, we employed the famous global attack strategy, Metattack \cite{zugner2018adversarial1}, to generate the adversarial perturbations. In particular, we set a relatively strong attack operation by injecting 25\% of noisy links in the experiments. All experiments take the average performance together with standard deviation under 10 independent runs. All defense methods are adopted from the default parameter settings. Moreover, the setting of two threshold parameters $t_1$ and $t_2$ are determined based on the grid search process in [0, 0.5] with the interval of 0.05 and [0.5, 0.9] with the interval of 0.1, respectively. Moreover, we selected the classic GCN layer as the backbone of all GNN layers in the proposed De2GNN. The degree bound for determining the tail nodes is set as 5, as in a previous work \cite{liu2020towards}. (The codes for implementing De2GNN will be publicly available at \href{https://github.com/alexfanjn/De2GNN}{https://github.com/alexfanjn/De2GNN}.)

	\subsection{Experimental Results and Discussion}

	\subsubsection{Accuracy of All/Tail Nodes without Attacks}
	
	We first analyze the classification performance on clean datasets without attacks for all nodes and tail nodes in the test set. From Table \ref{tab:acc_before}, we can observe that the proposed De2GNN can obtain a comparable classification performance to other baselines for both all nodes and tail nodes in the test set. De2GNN has a similar performance as the original GCN, and the best performance on Citeseer for both all nodes and tail nodes in the test set.

	\subsubsection{Accuracy of All/Tail Nodes with Attacks}
	We then investigate the classification performance of all nodes and tail nodes against adversarial attacks. As shown in Table \ref{tab:acc_after}, the proposed De2GNN obtains the best defense performance than all other baseline methods in all three datasets. Among all other baselines, GCN obtains the worst classification performance as it is designed without any defense. Models SVD, RGCN, and Median demonstrate a better defense performance than GCN. In addition, Jaccard shows the overall best defense performance of all methods except for the proposed De2GNN. 
	
	Besides the classification performance for all nodes, we also explore the classification performance of tail nodes (i.e., degree $\leq$ 5), which is the main objective of this work. As shown in Table \ref{tab:acc_after}, the proposed De2GNN also demonstrates a strong performance on tail nodes, especially for Cora and Pubmed datasets. The overall performance for tail nodes of all baselines follows a similar trend as their performance for all nodes. Generally speaking, the classification performance for tail nodes tends to be a little bit more difficult than all nodes, as the prior group only contains those nodes with relatively low degree.

	\begin{figure*}[t]
		\subfigure[\textbf{Cora}]{
			\begin{minipage}[]{0.33\linewidth}
				\includegraphics[scale=0.35]{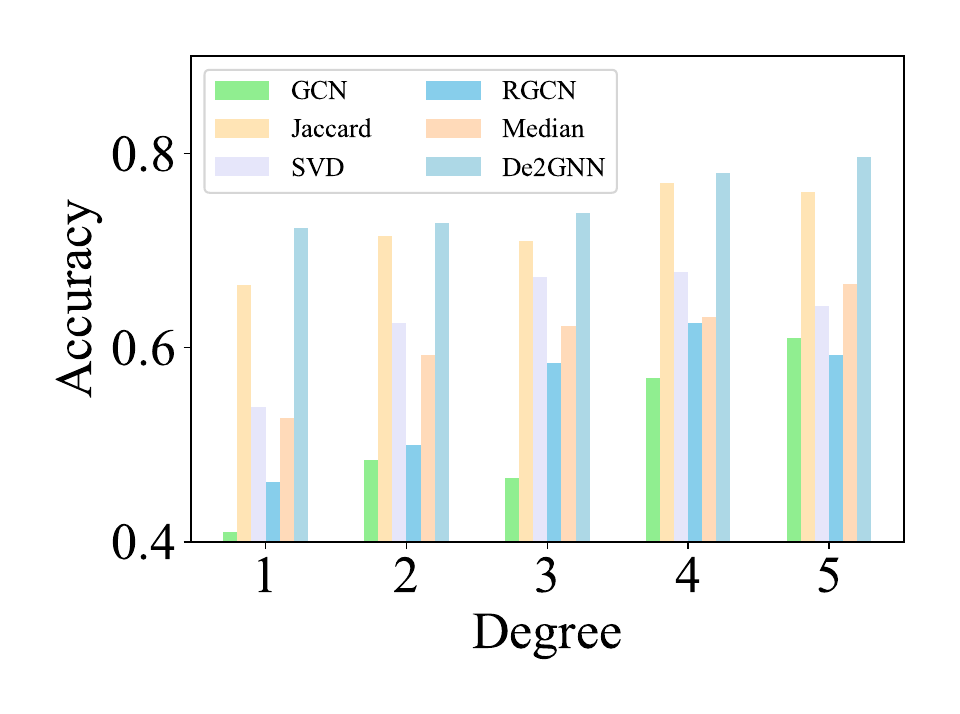}
			\end{minipage}%
		}%
		\subfigure[\textbf{Citeseer}]{
			\begin{minipage}[]{0.33\linewidth}
				\includegraphics[scale=0.35]{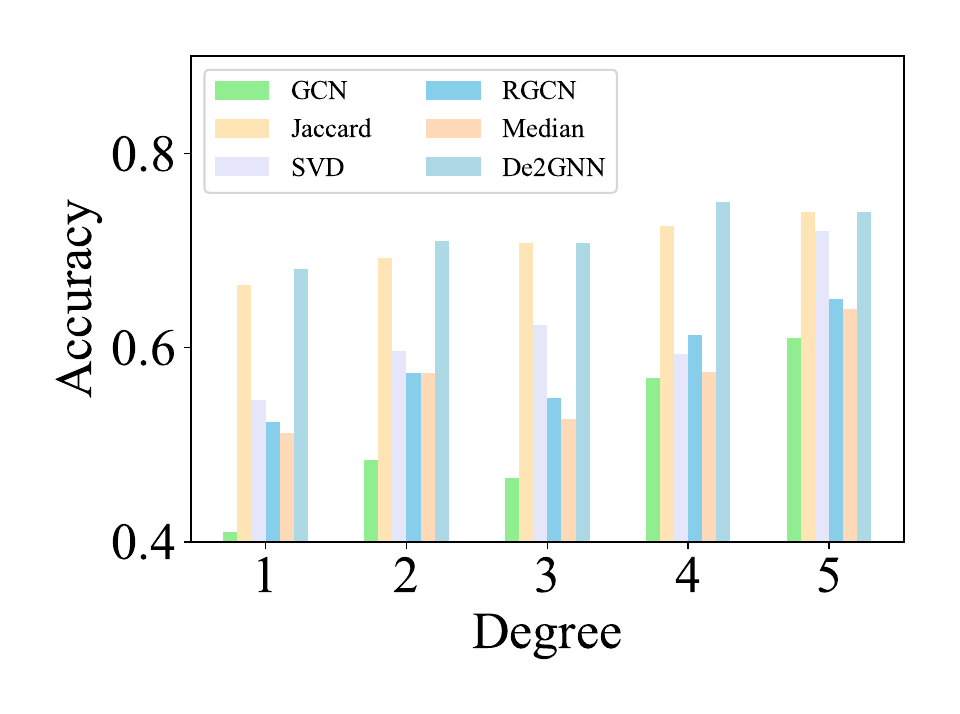}
			\end{minipage}%
		}%
		\subfigure[\textbf{Pubmed}]{
			\begin{minipage}[]{0.33\linewidth}
				\includegraphics[scale=0.35]{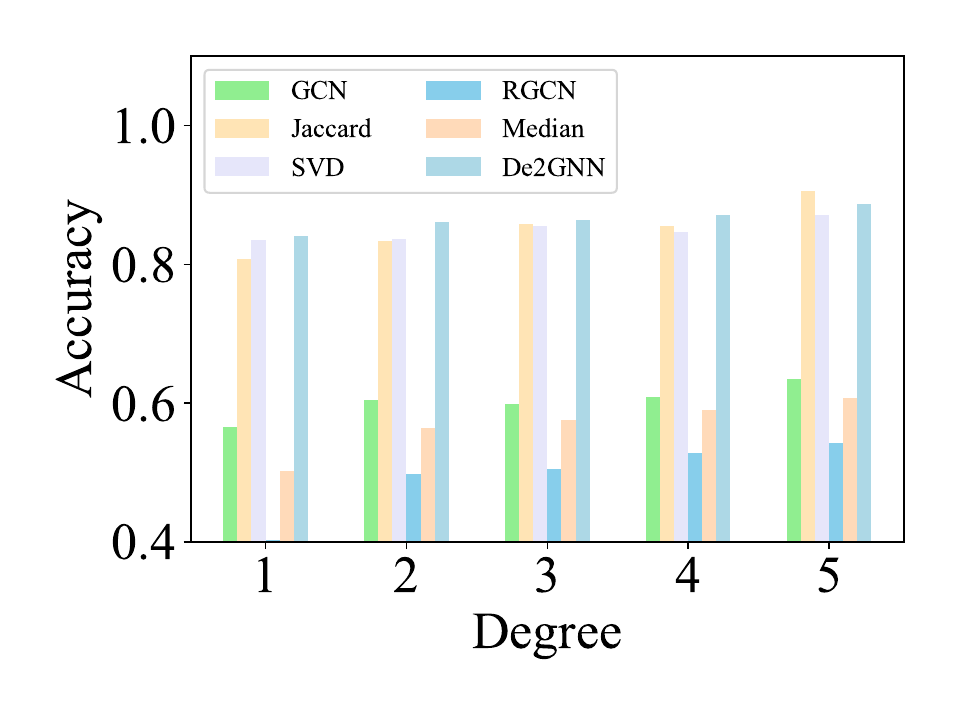}
			\end{minipage}%
		}%
		\centering
		\caption{Accuracy of baselines with regard to node degree on perturbed datasets by injecting 25\% of noisy links via Metattack.}
		\label{fig:acc_per_degree}
	\end{figure*}

	\subsubsection{Accuracy of Tail Nodes w.r.t Degree with Attacks}
	Further, we analyze the accuracy of baselines for tail nodes against adversarial attacks in a more fine-grained setting. We present the accuracy of all baselines with regard to the degree of nodes, as shown in Fig. \ref{fig:acc_per_degree}. We can observe that De2GNN achieves the best performance in almost all cases (except on Pubmed when the degree is 5), especially for those nodes with extremely low degree, indicating its superiority against adversarial attacks. Among all other baselines, Jaccard obtains the overall best performance. In general, all methods obtain a better performance with the increase of degree. Moreover, compared to other baselines, De2GNN tends to have a stable classification performance in different degrees, demonstrating its debiasing effect for nodes with different degrees.
	
	\subsubsection{Ablation Studies}
	
	From previous experiments, we know that the proposed De2GNN demonstrates good potential for mitigating the degree bias against adversarial attacks even under a strong noise. Therefore, in the following, we investigate the contribution of each module of De2GNN. Specifically, we analyze the four important modules of De2GNN, including the removal of heterophilic links (De2GNN$_{-{\rm hetero}}$), addition of homophilic links (De2GNN$_{-{\rm homo}}$), $k$NN augmented graph (De2GNN$_{-k{\rm NN}}$), and attention mechanism (De2GNN$_{-{\rm attn}}$), in terms of classification performance of tail nodes. For De2GNN$_{-{\rm hetero}}$ and De2GNN$_{-{\rm homo}}$, we remove the module of heterophilic link removal and homophilic link addition from the original De2GNN, respectively. For De2GNN$_{-k{\rm NN}}$, we remove the graph view of $k$NN augmented graph but retain the view of the hetero-homo augmented graph. For De2GNN$_{-{\rm attn}}$, we discard the node-wise attention mechanism. Instead, we directly employ a concatenation operation to combine the node representations obtained from the two graph views, together with an extra MLP (i.e., multi-layer perceptrons) network to learn the final outputs.
	
	The contribution of each module of De2GNN is shown in Table \ref{tab:aba}. As we can observe, each part contributes to the final performance of De2GNN. Among them, the model De2GNN$_{-{\rm hetero}}$ shows a significant performance drop, indicating the importance of removing heterophilic links in adversarial defense. This is reasonable as the graph has been injected with some strong noise. The addition of homophilic links also has a positive impact on the effectiveness of De2GNN. Moreover, if we remove the $k$NN augmented graph, De2GNN also demonstrates a slight performance drop. Last but not least, De2GNN$_{-{\rm attn}}$ shows a relatively low classification performance, indicating that the employed attention mechanism can help obtain adaptive representations from two different graph views.

	\begin{table}[]
		\centering
		\caption{Ablation study of each module of De2GNN in terms of accuracy of tail nodes (i.e., degree $\leq$ 5).}
		\resizebox{\linewidth}{!}{
			\begin{tabular}{l|c|c|c}
				\bottomrule
				\textbf{\rule{0in}{0.15in}Models} &
				\textbf{Cora} &
				\textbf{Citeseer} &
				\textbf{Pubmed} \\ \hline\hline
				
				De2GNN$_{-{\rm hetero}}$ & \rule{0in}{0.15in}0.6265\scriptsize{$\pm$0.0080} & 0.6146\scriptsize{$\pm$0.0132} & 0.7322\scriptsize{$\pm$0.0119} \\
				
				De2GNN$_{-{\rm homo}}$ & 0.7184\scriptsize{$\pm$0.0075} & 0.6919\scriptsize{$\pm$0.0055} & 0.8266\scriptsize{$\pm$0.0049} \\

				De2GNN$_{-k{\rm NN}}$ & 0.7480\scriptsize{$\pm$0.0087} & 0.7007\scriptsize{$\pm$0.0028} & 0.8376\scriptsize{$\pm$0.0041} \\

				De2GNN$_{-{\rm attn}}$ & 0.6971\scriptsize{$\pm$0.0112} & 0.6835\scriptsize{$\pm$0.0108} & 0.8242\scriptsize{$\pm$0.0186} \\

				De2GNN & \textbf{0.7491\scriptsize{$\pm$0.0101}} & \textbf{0.7082\scriptsize{$\pm$0.0036}} & \textbf{0.8487\scriptsize{$\pm$0.0035}} \\ 
				\toprule
		\end{tabular}}%
		\label{tab:aba}
	\end{table}

	\begin{figure}[]
		\centering
		\includegraphics[width=0.8\linewidth]{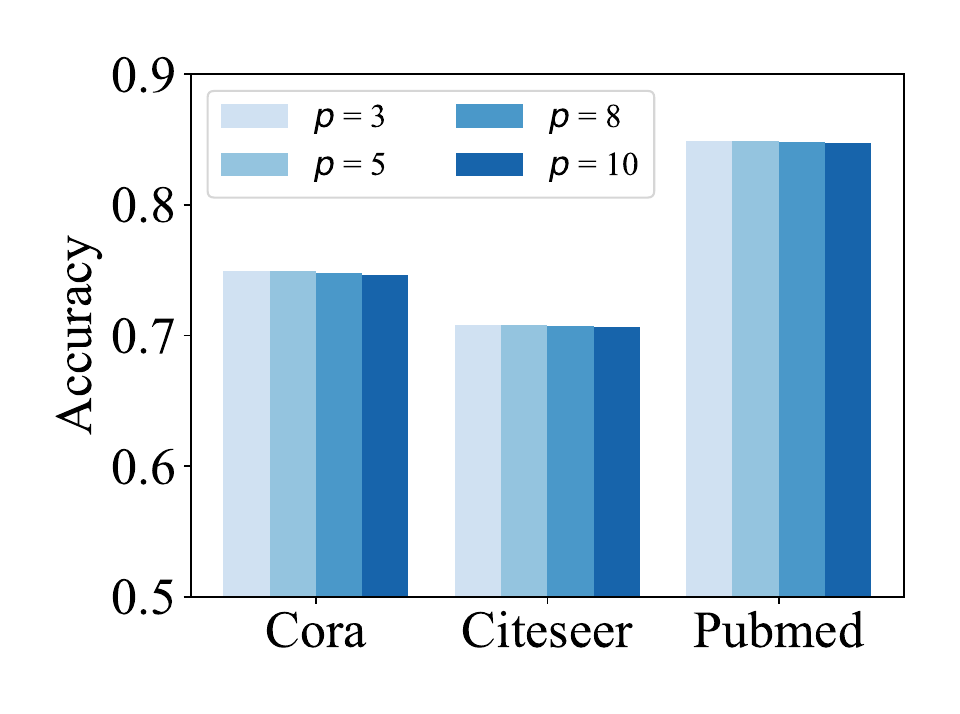} 
		\caption{Parameter analysis for the number of added nodes during the addition of homophilic neighbors for tail nodes under three datasets.}
		\label{fig:para_k}
	\end{figure}
	
	\begin{figure}[]
		\centering
		\includegraphics[width=0.8\linewidth]{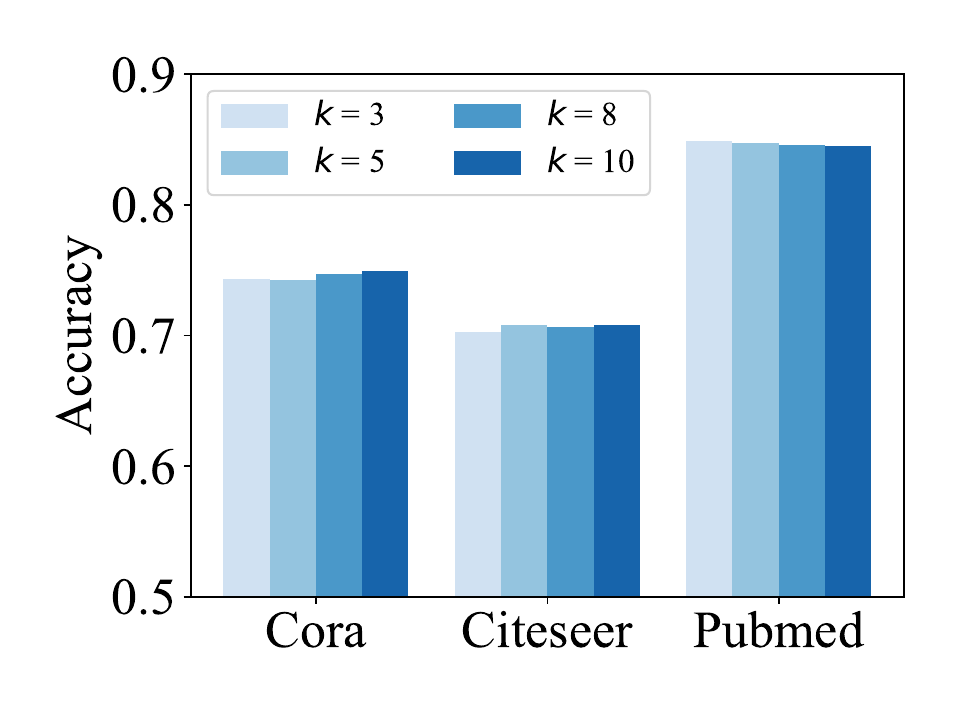} 
		\caption{Parameter analysis for the number of neighbors during the $k$NN augmented graph construction under three datasets.}
		\label{fig:para_p}
	\end{figure}
	
	\subsubsection{Parameter Analysis}
	Finally, we analyze the sensitivity of some important parameters in terms of the classification performance for tail nodes. The investigated parameters include the number of added neighbors during the addition of homophilic neighbors for tail nodes ($p$) and the number of neighbors during the $k$NN augmented graph construction ($k$). Specifically, for both $p$ and $k$, we set the same parameter ranging group as $\{3, 5, 8, 10\}$.
	
	From Fig. \ref{fig:para_p}, we can observe that, the performance of De2GNN tends to be stable during different settings of $p$ in all datasets. Particularly, a lower $p$ tends to obtain a slightly better prediction accuracy than a higher $p$, indicating that it is not always effective to add a large number of neighbors to tail nodes. The possible reason is that these homophilic neighbors are obtained from the surrogate GNN model which may contain some noisy predictions. For the number of added nodes during the addition of homophilic neighbors for tail nodes, we also observe a relatively stable classification performance with varied $k$ in all datasets. The above results support that the proposed De2GNN is a generalizable model as it is not sensitive to diverse parameter settings.

	\section{Conclusion}\label{sec:con}
	In this work, we investigate the structural bias issue of current graph neural network models in graph adversarial defense scenarios. We first verify the existence of structural bias that current defense models tend to have a worse prediction performance on nodes with lower degree than those nodes with higher degree through detailed analysis. To address this, we further propose a simple yet effective defense\&debiasing model, De2GNN, to defend against possible adversarial perturbations and mitigate the structural bias on nodes with low degree simultaneously. Specifically, the proposed De2GNN contains the hetero-homo augmented graph view which removes possible heterophilic links and adds homophilic links, and the $k$NN augmented graph view which is generated based on node features. Finally, a node-wise attention mechanism is employed to adaptively combine the information from the above two views. Experiments have been conducted to demonstrate the effectiveness of the proposed model in terms of defense and debiasing performance.

	\bibliographystyle{IEEEtran}
	\bibliography{IEEEabrv, mybib}
	
\end{document}